\documentclass[10pt,twocolumn,letterpaper]{article}

\usepackage{iccv}
\usepackage{times}
\usepackage{epsfig}
\usepackage{graphicx}
\usepackage{amsmath}
\usepackage{authblk}

\def\no{no\onedot}
\DeclareMathOperator*{\argmin}{\mathrm{argmin}}

\DeclareMathOperator*{\Cov}{\mathrm{Cov}}

\DeclareMathOperator*{\arctantwo}{\mathrm{arctan2}}
\usepackage{amssymb}

\usepackage{tikz}
\usetikzlibrary{calc,positioning}
\usepackage{pgfplots}
\usepackage{pgfplotstable}
\pgfplotsset{width=10cm,compat=1.14}

\usepackage[pagebackref=true,breaklinks=true,colorlinks,bookmarks=false]{hyperref}

\iccvfinalcopy 

\def\iccvPaperID{1317} 

\begin{document}

\title{Monocular 3D Object Detection and Box Fitting Trained End-to-End\\ Using Intersection-over-Union Loss\vspace{-1ex}}

\author[1, 2]{Eskil J\"orgensen}
\author[1]{Christopher Zach}
\author[1]{Fredrik Kahl\vspace{-1ex}}
\affil[1]{Chalmers University of Technology}
\affil[2]{Zenuity AB}
\affil[ ]{\texttt{\{eskilj, zach, fredrik.kahl\}@chalmers.se}\vspace{-2ex}}

\maketitle

\begin{abstract}
    Three-dimensional object detection from a single view is a challenging task which, if performed with good accuracy, is an important enabler of low-cost mobile robot perception.
    Previous approaches to this problem suffer either from an overly complex inference engine or from an insufficient detection accuracy.
    To deal with these issues, we present \textsc{SS3D}, a \underline{s}ingle-\underline{s}tage monocular \underline{3}D object \underline{d}etector. The framework consists of (i) a CNN, which outputs a redundant representation of each relevant object in the image with corresponding uncertainty estimates, and (ii) a 3D bounding box optimizer.
    We show how modeling heteroscedastic uncertainty improves performance upon our baseline, and furthermore, how back-propagation can be done through the optimizer in order to train the pipeline end-to-end for additional accuracy.
    Our method achieves SOTA accuracy on monocular 3D object detection, while running at 20 \emph{fps} in a straightforward implementation.
    We argue that the \textsc{SS3D} architecture provides a solid framework upon which high performing detection systems can be built, with autonomous driving being the main application in mind.
\end{abstract}


\section{Introduction}
The problem of detecting objects of interest and inferring their 3D properties is a central problem in computer vision with a plethora of applications. In the last decade, a rapid growth in research and development of mobile robotics and autonomous driving has established a pressing demand for accurate and efficient perception systems to enable safe and real-time machine operation among humans and other moving objects. Object detection plays a crucial role in such systems and it is the primary motivation behind this paper.

While recent advances in the field of 2D object detection have been important for the development of autonomous cars, there is still a need for improvement when it comes to upgrading the detection from the image plane to real world poses. This task, called \emph{3D object detection}, naturally lends itself to methods involving radar, lidar, stereo or other depth sensors. Modern lidar sensors, in particular, are a popular choice for depth and 3D, but there are two important reasons to consider methods only involving monocular vision. The first one is that of modularity and redundancy in the perceptual system (\eg in the case of sensor failures). The second one is the fact that, as of today, image sensors are still significantly cheaper than high precision lidars, and cameras are already built into an increasing number of personal cars.

\begin{figure}[t]
\centering
\includegraphics[trim={0cm 0cm 0cm 0.8cm},clip, width=0.475\textwidth]{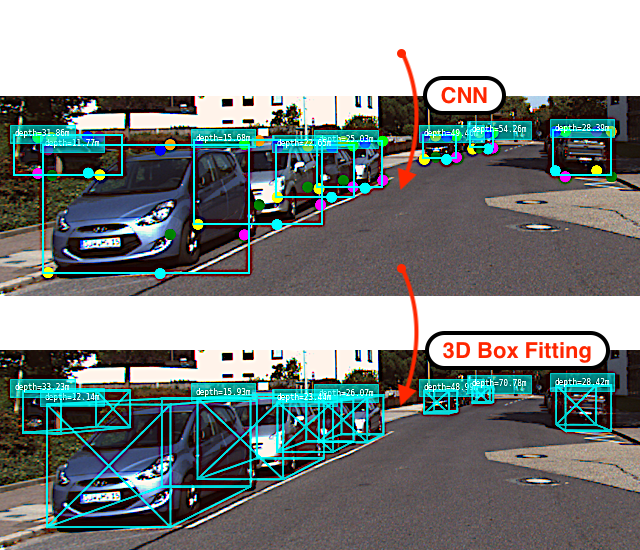}
\caption{In the \textsc{SS3D} pipeline, an image is fed to a fully-convolutional neural network, which performs 2D detection and regresses keypoint locations, orientation, dimensions and distance for each object. Using a learned weighting of these outputs, an optimizer yields accurate 3D bounding boxes.}
\label{fig:demo}
\end{figure}

Examining the literature on 3D object detection reveals that most methods based on purely vision are too slow for real-time applications and that the accuracy is far inferior to methods using other sensors such as lidars, see the \textsc{KITTI} benchmark~\cite{KITTI} for an overview. This is true for both methods based on monocular and stereoscopic vision, and it motivates us to improve the insufficient accuracy and overly complex processing steps of current state-of-the-art.

In this work, we address the problem with a pragmatic outlook. We recognize the overwhelming benefits of learning based approaches using convolutional neural networks for object detection. On the other hand, we note that deep neural networks still are the major computational bottlenecks for real-time applications even on specialized hardware. Therefore, we avoid processing the image multiple times, which is inspired by
single-stage detectors (SSD), e.g.~\cite{ssd,yolo9000}. More specifically, our framework \textsc{SS3D} works with a single-stage detection network and off-the-shelf
non-max suppression followed by a non-linear least squares optimizer that generates per-object canonical 3D bounding box parameters. By
decoupling CNN-based 2D predictions and optimization-based 3D fitting, the network is not required to learn the analytically known perspective projection of a pinhole camera. We also show that the whole pipeline can be trained end-to-end using a loss function based on the 3D intersection-over-union. Our method works in real-time and takes a single image as input, see Figure~\ref{fig:demo}.

Our main contributions are:
\begin{itemize}
    \item We propose an efficient fully convolutional single-stage network architecture for accurate monocular 3D object detection by regressing a surrogate 3D representation and using it for geometric fitting of 3D boxes.
    \item We demonstrate how incorporating the assumption of heteroscedastic noise yields an increased detection performance in addition to corresponding covariance matrices for the estimated 3D boxes.
    \item We demonstrate how the network and 3D box fitting can be trained end-to-end using \emph{backprop through optimization} and how this further improves performance.
\end{itemize}
Our approach achieves SOTA performance among all published monocular vision methods on the \textsc{KITTI} 3D object detection dataset while also running faster than previous methods.
  
\section{Related work}

In this section we highlight some recent research on 3D object detection and rigid 3D shape fitting.

\textbf{2D object detection} refers to the task of estimating axis aligned bounding boxes covering each detected object in the image, and possibly classifying them further into categories. The availability of large datasets containing bounding box annotations~\cite{Everingham15,mscoco} has enabled continuous improvements and led to highly accurate models able to handle many object classes.
Object detection methods are commonly subdivided into two-stage detectors \cite{fasterrcnn, fpn} and single-stage detectors \cite{yolo9000, ssd, focalloss, squeezedet}. Two-stage detectors use an initial stage (often a large base network) to propose regions of interest and a subsequent approach (i.e.\ a DNN) to process each of these regions for a refined classification and bounding box regression. Single-stage detectors on the other hand use a single architecture to output the final bounding boxes and classification scores. They are typically significantly faster than two-stage detectors at the expense of a reduced accuracy.

\textbf{2D object detection with rigid pose estimation} assumes that each object category has a fixed and known shape, which enables the estimations of a 3D object pose from a single image via 2D-3D correspondences.
Prior works include~\cite{bb8, rtssd6d}, which estimate the coordinates of the projected 3D bounding box corners.

\textbf{3D object detection from monocular vision} estimates both the object's pose and size, and therefore faces additional challenges due to the intrinsic scale ambiguity of monocular reconstruction.
Xiang \etal \cite{3DVP} propose to cluster voxel representations extracted from a 3D database into \emph{voxel pattens}.
At inference time, 2D object detection predicts bounding boxes and classification scores for each of these voxel pattern, from which 3D information can be extracted. This framework was later extended to leverage CNNs \cite{SubCNN}.
Chen \etal \cite{Mono3D} utilize an energy function incorporating provided segmentation masks and location priors to prune a candidate set of 3D object predictions, and a final CNN stage predicts classification scores, 2D bounding boxes and orientation.
A 3-stage pipeline is proposed by Chabot \etal \cite{DeepMANTA}, where 2D bounding box and keypoint prediction steps are followed by a 3D similarity scoring stage. The 3D model database is augmented with additional 3D keypoints, and the final object pose is estimated from 2D-3D keypoint correspondences.

Mousavian \etal \cite{Deep3DBox} use pre-existing state-of-the-art two-stage detectors to extract object patches and use these as input for a DNN to estimate the object's orientation and size.
These predictions are used to create 3D bounding boxes, whose image projections fit tightly inside the respective 2D bounding boxes. 

Xu and Chen~\cite{MonoFusion} combine CNN-based single image depth estimation with a 2D proposal-based architecture to perform 3D object detection.
MonoGRNet~\cite{monogrnet} leverages several subnetworks corresponding to intermediate 3D regression tasks, and combines these to estimate the final 3D bounding box.

In comparison to the above approaches, we have a relatively simple (and hence fast) single-stage network. Further, we choose to combine multiple prediction tasks and focus on the proper weighting of these. We also show that it is feasible to directly train all the parameters end-to-end using a 3D intersection-over-union loss.

\textbf{3D object detection from other sensors} usually avoids the main challenges of accurate depth estimation by leveraging a stereo setup~\cite{3DOP}, lidar sensors~\cite{voxelnet, SECOND, PIXOR} or a combination of RGB and lidar sensors~\cite{frustum, MV3D, avod, pointfusion, posercnn, Liang_2018_ECCV}. We leave the extension of our method to such setups as future work.

For our \textbf{end-to-end training} approach we leverage backpropagation through optimization. If the underlying optimization method is iterative, then these iterations are often unrolled to enable end-to-end training. Some other works rely on finite differences \cite{3DRCNN_CVPR18}, which quickly becomes intractable as the problem increases in dimension. Notable exceptions, where exact gradients are propagated through an application-dependent optimizer, include~\cite{ionescu2015matrix} and~\cite{barron2016fast}.

\section{Problem formulation and method overview}
We address the problem of monocular 3D object detection, which we more formally define as follows: report all objects of interest in a given color image $\mathbf x \in \mathbb R^{H\times W \times 3}$ and output for each object
\begin{itemize}
    \item its class label $\mathit{cls}$,
    \item the 2D bounding box represented by its top-left and bottom-right corners $\mathbf c = (x_1, y_1, x_2, y_2)$,
    \item and its 3D bounding box encoded as 7-tuple $\mathbf b = (h, w, l, x_c, y_c, z_c, \theta)$, where $(h,w,l)$ are the spatial dimensions, $(x_c, y_c, z_c)$ is the center point (in the camera coordinate frame), and $\theta$ is the yaw angle of the 3D box.
\end{itemize}
As commonly done in vehicle-oriented applications, we model object orientation only with the yaw angle, $\theta$, setting roll and pitch to zero.

\begin{figure*}
     \centering
    \includegraphics[trim={.2cm .7cm 0.1cm .6cm},clip, width=1.0\textwidth]{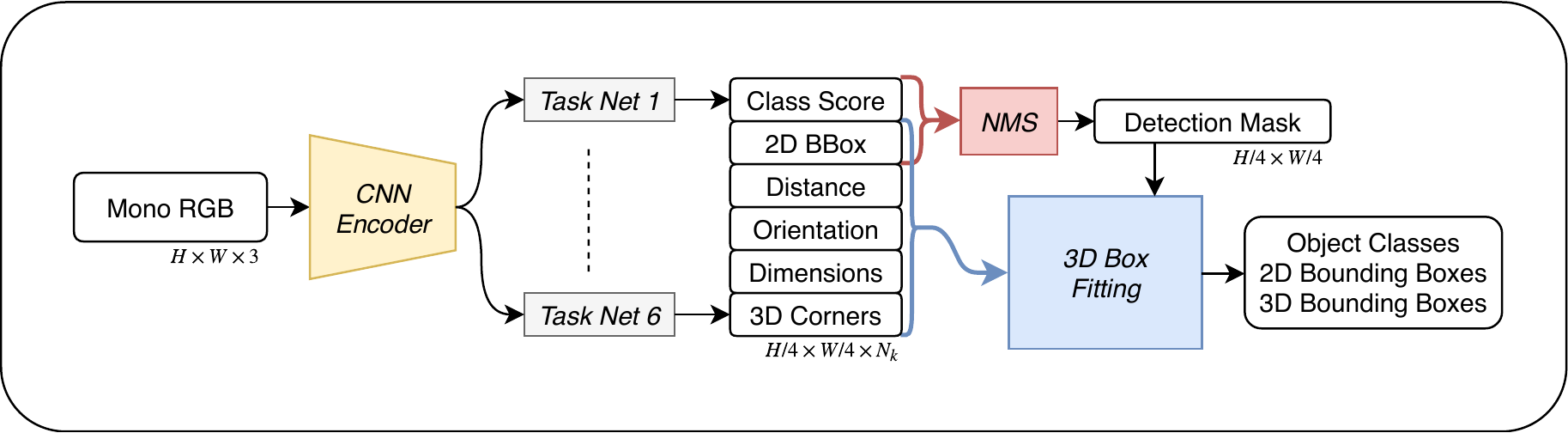}
     \caption{An overview of the \textsc{SS3D} pipeline.}
     \label{fig:ss3d-mono}
\end{figure*}

\paragraph{Method overview. }
The proposed \textsc{SS3D} network architecture is illustrated in Figure~\ref{fig:ss3d-mono}.
The pipeline consists of the following steps:
\begin{enumerate}
\item a CNN performs object detection (yielding class scores) and regresses a
  set of intermediate values used later for 3D bounding box fitting,
\item non-maximum suppression is applied to discard redundant detections,
\item and finally the 3D bounding boxes are fitted given the intermediate predictions using a non-linear least-squares method.
\end{enumerate}
One important aspect of our proposed method is efficiency: the involved CNN is
relatively light-weight and the fitting of bounding boxes for each detected
object is performed independently and in parallel. Further, the input image is accessed only once in the 
CNN-based feature extraction stage.

\subsection{Surrogate regression targets}

The training data is given as ground truth annotations ($\mathit{cls}_{GT}$, $\mathbf c_{GT}$ and $\mathbf b_{GT}$) for each of the training images.
While 2D bounding box parameters $\mathbf c$ are regressed directly, we assume that 3D bounding boxes $\mathbf b$ are not suitable for direct CNN-based prediction.
The reason is, that 3D box parameters are rather abstract 3D quantities not directly linked with how the object appears in the image.
Further, the loss function (and thus the loss gradient) is more informative due to the higher dimensional regression targets (26 instead of 7 dimensions).
We therefore aim to ease the regression task for the network by reparametrizing the bounding box parameters to proxy or surrogate targets $\{f_k(\mathbf b)\}_{k=1}^{26}$ (defined in Section~\ref{subsec:outputs}), which are quantities more closely related to the 2D image domain.
These proxy targets provide sufficient information to extract the 3D box parameters $\mathbf b$.

\subsection{Object detection}
Classification probabilities (class scores) are obtained from a soft-max layer, and are subsequently fed (together with the 2D bounding box estimates $\mathbf c$) to a standard non-maximum suppression (NMS) function. Therefore, SS3D contains a single-stage standard object detector and could be trained on additional 2D object detection data.

Detection is done for each class separately using its class score map from the soft-max outputs and the 2D bounding box map. Pixels with class score less than 0.7 are rejected. From the remaining candidates, a greedy NMS algorithm is run to select as detections all pixels which do not have $\mathit{IoU}_{2D}>0.3$ with any other candidate of a higher class score. The regression values
\begin{equation}
    \mathbf y = \hat {\mathbf f}(\mathbf b) \in \mathbb R^{26}
\end{equation}
corresponding to these detected pixels are then used to fit 3D bounding boxes as outlined in the next section.

\subsection{3D bounding box fitting}
For each detection, the network regression output vector $\mathbf y$ is used to estimate 3D bounding box parameters $\mathbf {\hat b} = \argmin_{\mathbf b} E(\mathbf{b};\mathbf{y})$, where
\begin{equation}
    \label{eq:cost}
    E(\mathbf b;\mathbf{y}) = \sum_{i=1}^{26} r_i(\mathbf b;\mathbf{y})^2 = \sum_{i=1}^{26} \left(w_i(y_i - f_i(\mathbf b))\right)^2
\end{equation}
is a non-linear least-squares objective, which also contains confidence weights $w_i$ discussed below.

This weighted non-linear least squares problem can conveniently be solved in parallel for each detection.
The objective in (\ref{eq:cost}) may have a number of spurious local minima, but nevertheless, a good initial solution can be obtained by ``piece-wise'' fitting of parameters described in Section~\ref{subsec:box_initialization}.

The weights, $w_i$, should be set to account for the fact that the different regression values are expressed in different units and have different associated uncertainties. Assuming zero-mean Gaussian aleatoric regression uncertainties the weights should ideally be set to $w_i = 1/\sigma_i$, where $\sigma_i^2$ is the respective variance, which turns (\ref{eq:cost}) into maximum likelihood estimation.
This allows the covariance matrix of the estimated bounding box to be expressed as
\begin{equation}
    \Cov(\mathbf{\hat b}) = \left(\nabla_{\mathbf{b}}^2E(\mathbf b;\mathbf{y})\right)^{-1}
    ,
\end{equation}
making our framework particularly useful \eg for an object tracking algorithm. It would not be as straightforward to obtain similar uncertainties from methods which rely only on black-box models for estimation of the 3D box parameters \cite{MonoFusion}. 
We compare three different methods to set the weights, $w_i$, explained below in Section~\ref{sec:train}.
\section{Training approach}
\label{sec:train}

The network weights $\mathbf W$ are learned using loss functions described in this section. Additionally, the confidence weights $\mathbf{w}$ from (\ref{eq:cost}) have to be learned. We study three different methods of learning $\mathbf W$ and $\mathbf w$ jointly.
All methods use pixel-wise cross entropy as classification loss, $\mathcal L_{cls}$. The loss functions for the regression tasks on the other hand are altered between the methods as described below. In each case, $w_i$ is extracted as the inverse of $\sigma_i$.

\subsection{Method 1: Homoscedastic noise}
The traditional assumption is that the statistical noise level is constant for the complete dataset, but the parameters of the noise model are unknown and have to be estimated from training data via maximum likelihood estimates. Since we have vectorial regression outputs $\mathbf y$, leveraging a standard zero-mean Gaussian assumption on the noise model means that a full ($26\times 26$) covariance matrix has to be estimated.
We restrict the covariance matrix to be purely diagonal for efficiency, and obtain the following combined loss function (similar to~\cite{kendall2017multi}),
\begin{align}
    \mathcal L_1 &= \mathcal L_{cls} \nonumber \\
                 &+ \frac{1}{N} \sum_{j=1}^N \sum_{k=1}^{26} \left( \frac{\big\| \mathbf y_k^{(j)}-f_k(\mathbf b^{(j)}) \big\|^2}{2\sigma_k^2} + \log \sigma_k \right).
\end{align}
The variances $\sigma_k$ are trained jointly with the network parameters $\mathbf W$. Here
$k$ ranges over the elements of the regression output and $j$ refers to the index of training samples.

In principle one could introduce another level of relative weighting between the classification loss $\mathcal L_{cls}$ and the regression losses.
Introducing this extra set of weights has little impact due to the following reasons:
(i) each task has its own relatively ``flexible'' head, (ii) the common encoder is pretrained on ImageNet and therefore features significant inertia, and (iii) we use the Adam optimizer with adaptive step sizes.
Thus, the joint loss above largely decouples during early stages of training, and we do not consider relative weights in the above and subsequent loss functions.

\subsection{Method 2: Heteroscedastic noise}

The assumption of homoscedastic noise uniformly affecting the network prediction may be unrealistic.
Sample-specific effects such as viewing angles and occlusions may lead to non-homogeneous effects in the regressed output provided by the network.
More concretely, suppose that an object is heavily occluded such that only a small part of it is visible. While it might not be possible to infer its dimensions with high confidence, it is still possible to form a posterior estimate of its dimensions learned from the dataset, albeit with a higher variance.
Ideally, the network is able to regress the variance (or precision) of the predicted output in order to downweight unreliable predictions in order to pass this information to the bounding box estimation stage.
In this setting of heteroscedatic noise each input has its own noise model, which---as above---we assume to be a zero mean Gaussian with diagonal covariance matrices.
In order to predict the variances on unseen data, the network output is extended to provide the respective variances in addition to the regression outputs, as done in \cite{kendall2017what}.
The combined loss function now reads as
\begin{align}
  \mathcal L_2 &= \mathcal L_{cls} \nonumber \\
               &+ \frac{1}{N} \sum_i \left( \frac{(y_i-f_i(\mathbf b_i)^2)}{2\sigma_i^2} + \log\sigma_i - \log P_{\text{prior}}(\sigma_i^{-2}) \right).
                 \label{eq:method2_loss}
\end{align}
The index $i$ ranges now over all output elements for all pixels in the support regions of objects for the entire training set. Support regions are image regions ensured to observe a particular object and are explained in detail in Section~\ref{subsec:outputs}.

Hence, $i=1,\dotsc, 26 \cdot \sum_{j=1}^N \operatorname{card}(\text{support region}_j)$.
Therefore $\sigma_i$ is only estimated from a single sample and we need to introduce a prior on the variances to avoid degeneracies if the regression error approaches 0.
A common prior is a Gamma distribution (with parameters $\alpha$ and $\beta$) on the inferred precision $\sigma_i^{-2}$.
If we choose $(\alpha, \beta)=(1,1/2)$ (i.e.\ the prior is an exponential distribution), then the above loss function is highly connected to a robust Cauchy M-estimator:
under the assumption that the network predicting $\sigma_i$ has infinite capacity, then $\sigma_i$ can be minimized out, leading to
\begin{align}
  \sigma_i^2 = 1 + (y_i- f_i(\mathbf b_i))^2,
\end{align}
whose reciprocal (the corresponding precision) can be identified as the weight function of the Cauchy M-estimator. 
Thus, the second term in (\ref{eq:method2_loss}) can be seen as robustified least-squares training error.

\subsection{Method 3: Backprop through optimization}
In the third approach, the box fitting weights $\mathbf w$ and network weights $\mathbf W$ were trained jointly using an end-to-end 3D intersection-over-union (IoU) loss. The loss is defined as
\begin{equation}
    \mathcal L_{IoU} = 1 -  \frac1{|\mathcal D|}\sum_{d\in\mathcal D} \max_{g\in \mathcal G}\left(\mathrm{IoU}_{\text{3D}}(\mathbf {\hat b}_d, \mathbf b_{g})\right),
\end{equation}
where $\mathcal D$ and $\mathcal G$ are the detections and ground truth boxes respectively, and differentiated as
\begin{equation}
    \frac{\partial \mathcal L_{IoU}}{\partial \mathbf W} =
    \sum_{d\in\mathcal D}
    \frac{\partial \mathcal L_{IoU}}{\partial \mathbf {\hat b}_d}
    \left( \frac{\partial \mathbf {\hat b}_d}{\partial \mathbf y_d} \frac{\partial \mathbf y_d}{\partial \mathbf W} + \frac{\partial \mathbf {\hat b}_d}{\partial \boldsymbol\sigma_d} \frac{\partial \boldsymbol\sigma_d}{\partial \mathbf W} \right).
\end{equation}
Differentiating the box estimator to find the Jacobians $\partial \mathbf b / \partial \mathbf y$ and $\partial \mathbf b / \partial \boldsymbol \sigma$ can be done through the implicit function theorem. Given network outputs, $(\mathbf y, \boldsymbol \sigma)$, the optimal bounding box $\mathbf{\hat b} = \argmin_{\mathbf b} E(\mathbf{b}, \mathbf{y}, \boldsymbol{\sigma})$ will satisfy
\begin{equation}
    \nabla_{\mathbf b} E(\mathbf{b}, \mathbf{y}, \boldsymbol{\sigma}) \rvert_{\mathbf b=\mathbf{\hat b}} = \mathbf 0,
\end{equation}
which in a neighborhood around the optimum defines a mapping
\begin{equation}
    \mathbf{\hat b}: (\mathbf y,\boldsymbol\sigma) \mapsto \argmin_{\mathbf b} E(\mathbf{b}, \mathbf y,\boldsymbol\sigma)
\end{equation}
with Jacobian given by
\begin{equation}
    \frac{\partial \mathbf {\hat b}}{\partial \mathbf y} = - \left[\frac{\partial^2E}{\partial \mathbf {\hat b}^2}\right]^{-1}
    \left[\frac{\partial^2 E}{\partial \mathbf y \partial \mathbf {\hat b}}\right]
    \approx - \left[\frac{\partial \mathbf r}{\partial \mathbf {\hat b}}\right]^+ \left[\frac{\partial \mathbf r}{\partial \mathbf y}\right]
\end{equation}
and equivalently for $\boldsymbol\sigma$. We can now define the total training loss as
\begin{equation}
    \mathcal L_3 = \mathcal L_{cls} + \lambda \mathcal L_{IoU}.
\end{equation}
Training with $\mathcal L_{IoU}$ is preferably done as fine-tuning on a model pre-trained using Method 2, since a non-zero gradient requires a non-zero intersection to begin with. The gradients are pushed only through the output pixels selected by the non-learnable NMS, but the training is end-to-end in the sense that $\mathbf W$ and $\mathbf w$ are optimized for 3D IoU directly.

\section{Network architecture}

This section provides details on the CNN architecture and formally introduces the
proxy regression targets.

\subsection{Network structure}
The trainable part of SS3D consists of a \emph{CNN Encoder} and a set of \emph{Task Nets}, depicted as yellow and grey boxes in Figure~\ref{fig:ss3d-mono}.
We use residual networks~\cite{resnet} stripped of the last fully connected layers for the encoder part. We found that ResNet34 works well, but the \texttt{drn\_c\_26} version of the open-source dilated residual network~\cite{DRN} turned out to yield better performance (at a slightly increased processing time) and was therefore used in the experiments. With the last two fully connected layers removed, the encoder effectively downsamples the input image 8 times, producing relatively dense feature vectors of dimension 512.

\begin{figure}[t]
\begin{center}
   \includegraphics[trim={0.3cm 0.9cm 0.9cm 0.8cm},clip, width=0.45\textwidth]{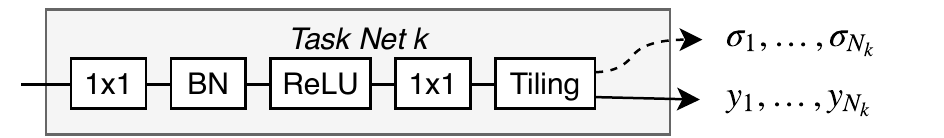}
\end{center}
   \caption{The task nets consist of two $1\times1$ convolution layers and apply a tiling upsampling. For two of our proposed methods, we let the regression task nets output an uncertainty $\sigma_i$ for each value $y_i$ at each output pixel.}
     \label{fig:task-net}
\end{figure}

The encoder output features are fed into 6 separate task networks, each consisting of two $1\times1$ convolution layers, separated by BatchNorm and ReLU as illustrated in Figure~\ref{fig:task-net}. A final tiling upsampling operation is used to obtain the final output resolution corresponding to one fourth of the input resolution. This design is adopted from \cite{thesisnet}. The stack of task nets could equivalently be interpreted and implemented as sixfold depth-wise separable convolutions with separate loss functions. The encoder is trained jointly with the task nets.

\subsection{Network outputs}
\label{subsec:outputs}
\begin{figure}[t]
\begin{center}
   \includegraphics[trim={0.2cm 0.7cm 0 .5cm},clip, width=0.35\textwidth]{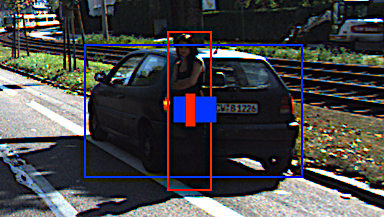}
\end{center}
   \caption{In the center of each object's ground truth 2D bounding box a rectangular \emph{support region} is created, with width and height set to 20\% of the former. Each output pixel in the support region holds regression targets and one-hot classification targets for the network. In the rare case when support regions overlap, the closer object is favored.}
     \label{fig:gt-mask}
\end{figure}

Some object detection networks \cite{fasterrcnn, yolo9000} have a relatively low resolution of output pixels (also called \emph{anchor points}) and assign each ground truth object to only one such anchor point. We take the approach used in~\cite{thesisnet}, which employs a higher output resolution, and connect each object with all output pixels in a rectangular \emph{support region} in the center of the object's bounding box (see Figure~\ref{fig:gt-mask}). The size of the support region is chosen in order to balance false negatives and false positives.
This output representation is in-between fully structured prediction networks (\eg~frequently used for semantic segmentation) and sparse outputs of pure object detectors.
During network training the regression loss is calculated over the support regions, while all other output pixels are considered as a don't-care-region. For the classification loss the areas outside all bounding boxes are treated as the background class, while the support region has the object class label (and all other output pixels are ignored).

The \textbf{classification} task head outputs one layer of probability logits for each object category of interest plus one layer for the background class. We use the following semantic foreground categories, \texttt{Cars}, \texttt{Pedestrians} and \texttt{Cyclists}, hence the output dimensions are $H/4 \times W/4 \times 4$. During inference, the class logits are fed pixel-wise to a SoftMax layer.

The regression task amounts to predicting the following proxy targets for given ground truth 3D box parameters $\mathbf{b}$:\\
\textbf{2D bounding boxes} coordinates $\mathbf c = (x_1, y_1, x_2, y_2)$ are learned in relative coordinates such that for each pixel $\mathbf p = (x, y)$ in the support region, the value
\begin{equation}
  f_{1:4}(\mathbf{c}) := \Delta\mathbf c = (x-x_1,\, y-y_1,\, x_2-x,\, y_2-y)
\end{equation}
is estimated. If $\mathbf c_{GT}$ is not available, it can be approximated as the rectangular envelope of the projected 3D box. This target is denoted by $f_{1:4}(\mathbf{b})$ and is used \eg in (\ref{eq:cost}).\\
The \textbf{distance}, $d$, is measured from the camera center to the object center, 
\begin{equation}
  f_5(\mathbf b) := d 
      = ||(x_c, y_c, z_c)||_2.
\end{equation}
While it might be sensible to predict $z_c$ directly, we saw only non-significant performance differences and chose to use $d$.\\
The \textbf{orientation}, $\theta$, defined as the object yaw angle in camera coordinates, is not easy to extract from the appearance of an object locally in an image, see discussions in \cite{Deep3DBox}. Instead, the network predicts the \emph{observation angle}, $\alpha$, encoded as
\begin{equation}
    f_{6:7}(\mathbf b) := (\sin(\alpha), \cos(\alpha)),
\end{equation}
where $\alpha := \theta - \arctantwo(x_c, z_c)$ is the angle of incidence on the target object (\ie it encodes from which angle the camera is looking at the object).
Encoding $\alpha$ can be done in various ways, and we use $\sin(\alpha)$ and $\cos(\alpha)$ as targets.
The binning method used in \cite{frustum,Deep3DBox} with joint classification and regression is effective but less suitable for integration in our end-to-end training approach.\\
For the object \textbf{dimensions}, $(h, w, l)$, we use targets $f_{8:10}(\mathbf b) := (\log h,\, \log w,\, \log l)$ similarly to~\cite{voxelnet,3DOP}.
This solves issues with negative sizes and states the loss in terms of relative rather than absolute errors.\\
\textbf{3D corners} are projected from the ground truth 3D bounding boxes into the image plane as
\begin{equation}
    \mathbf p_j = \frac P \lambda 
    \left(
        \begin{bmatrix}
            R_y(\theta) \\
            \mathbf 0
        \end{bmatrix}
        \begin{bmatrix}
            h & 0 & 0 \\
            0 & w & 0 \\
            0 & 0 & l
        \end{bmatrix}
        \frac{\mathbf v_j}2
        +
        \begin{bmatrix} x_c \\ y_c \\ z_c \\ 1 \end{bmatrix}
    \right)
\end{equation}
for each unit cube corner $\mathbf v_j \in \{-1,1\}^3$ where $P$ is the camera matrix, $\lambda$ the depth and $R_y(\theta)$ a yaw rotation matrix. With 8 corners and 2 image coordinates, there are 16 values in total to be estimated. The pixel locations are again encoded in relative coordinates, \ie
\begin{equation}
    f_{11:26}(\mathbf b) := \Delta \mathbf p_j = (x_{\mathbf p_j}-x, y_{\mathbf p_j}-y)
\end{equation}
is the target in the pixel $(x,y)$ of the support region.
This choice of proxy target is similar to \eg~\cite{bb8}, where pixel locations of ordered cuboid corners also are predicted.
In our case however, the object dimensions can differ within an object class.

\subsection{Initial 3D box estimate}
\label{subsec:box_initialization}

The non-linear least squares objective (\ref{eq:cost}) requires a
sufficiently good initial estimate $\mathbf{b}^0$ in order to avoid spurios local minima.
Given a prediction $\mathbf{y} \in\mathbb{R}^{26}$, one can first extract $\mathbf{c}$, $d$, $\alpha$, $\{\mathbf{p}_j\}$ and $(\log h, \log w, \log l)$.
The image coordinate of the object center is estimated as $u_x:= (x_1+x_2)/2$ and
$u_y:=(y_1+y_2)/2$, and the initial 3D object center is
$(x_c,y_c,z_c)^T = d (u_x,u_y,1)^T/\|(u_x,u_y,1)\|$. The yaw angle
$\theta$ is subsequently given by $\theta = \alpha + \arctantwo(x_c, z_c)$ and the object dimensions are extracted directly from the log-estimates.
\section{Experiments}

The proposed methods are evaluated primarily on the \textsc{KITTI} object detection benchmark~\cite{KITTI}.
This dataset consists of a number of image sequences with annotated 2D and 3D bounding boxes for a small number of object categories. We focus on three categories most relevant for autonomous vehicle applications in our evaluation: \texttt{Cars}, \texttt{Pedestrians} and \texttt{Cyclists}. 

\subsection{Learned uncertainties}
\begin{figure}
\begin{center}
   \includegraphics[trim={0.0cm 0.0cm 0 .0cm},clip, width=0.45\textwidth]{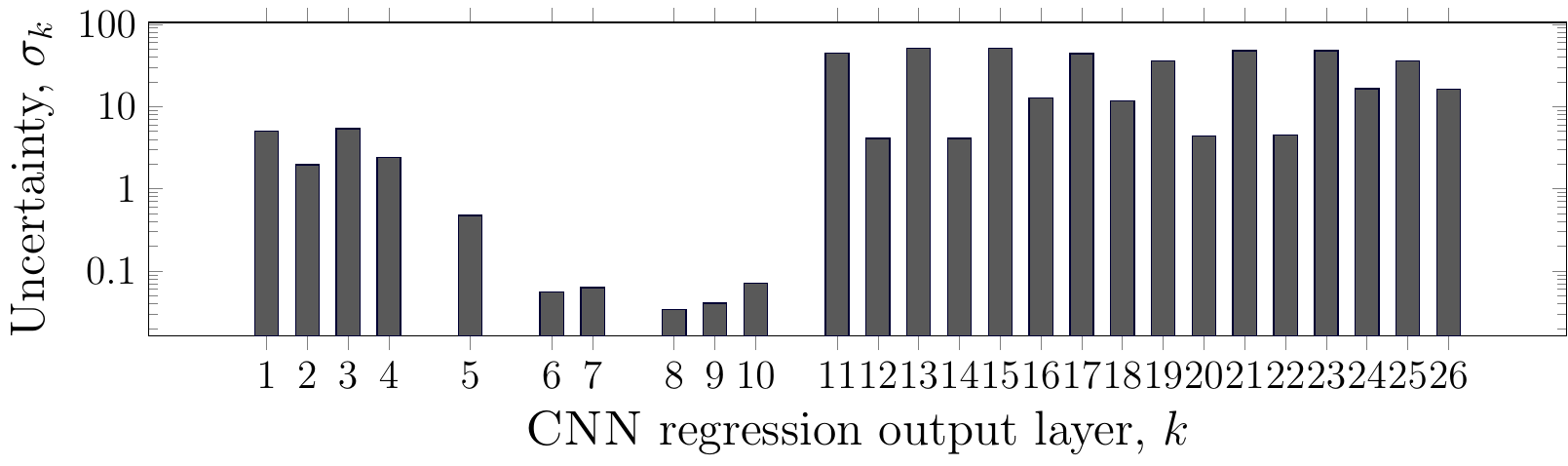}
\end{center}
   \caption{Learned uncertainties $\sigma_k$ when training Method~1 only on the \textsc{KITTI} car class for: 2D bounding box, distance, orientation, dimensions and 3D corners.}
     \label{fig:sigma_bar}
\end{figure}

\begin{figure}
    \begin{minipage}[b]{.49\linewidth} 
        \centering
        \includegraphics[width=0.95\textwidth]{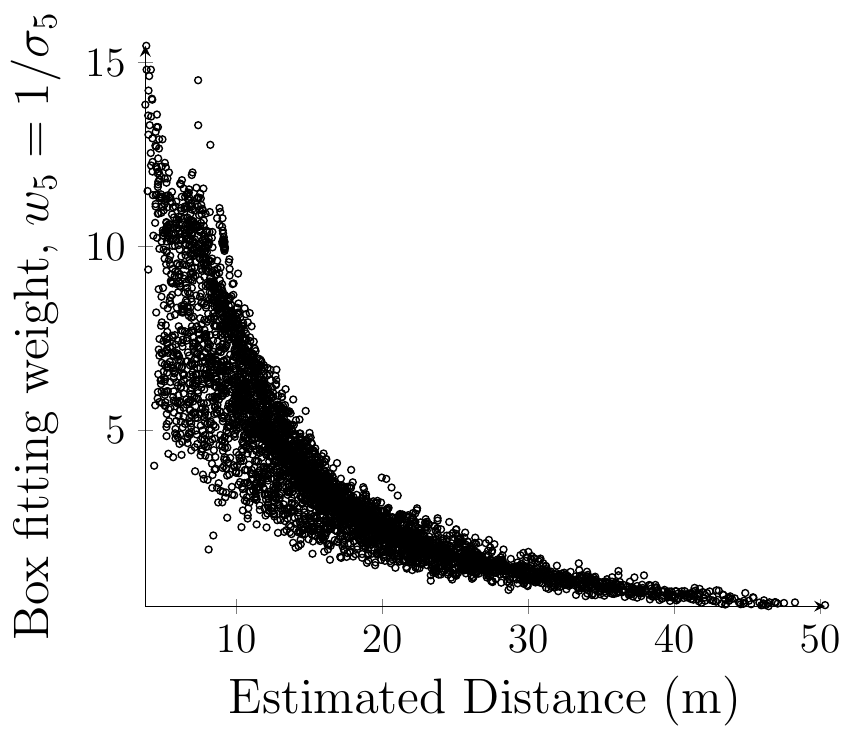}
        \label{fig:scatter_a}
    \end{minipage}
    \begin{minipage}[b]{.49\linewidth} 
        \centering
        \includegraphics[width=0.95\textwidth]{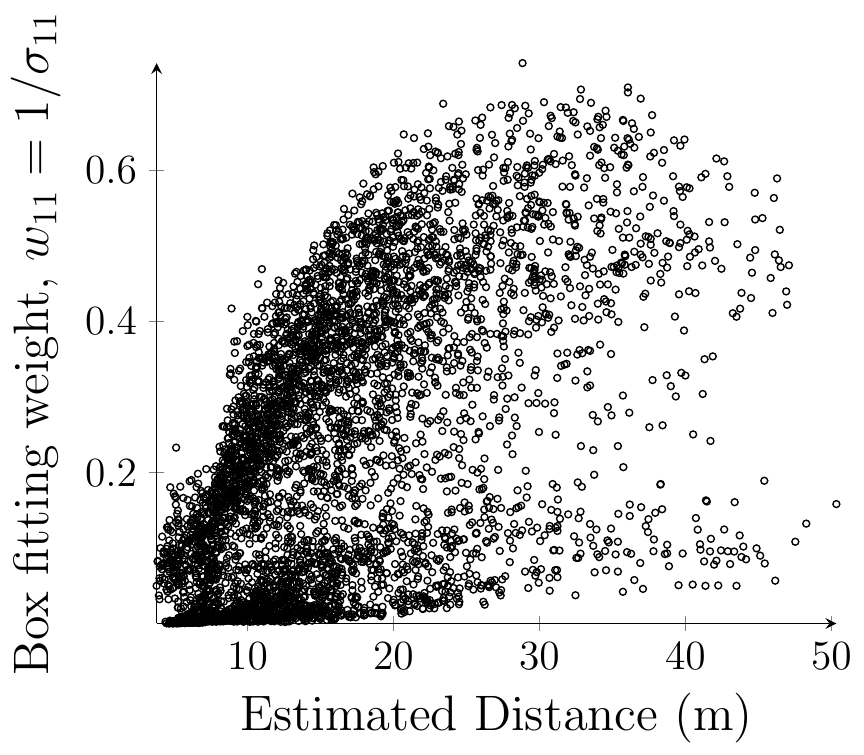}
        \label{fig:scatter_b}
    \end{minipage}
    \caption{Box fitting weights for Method~2 of the 5000 most confident detections plotted against estimated object distance. Left: object distance. Right: x-coordinate of projected top-left-front corner.}
    \label{fig:scatter}
\end{figure}

When training using a homoscedastic noise model, one expects the learned variances to reflect the scale differences caused by the different units for the regression targets. Figure \ref{fig:sigma_bar} validates that this is indeed the case.
In particular, image-space targets given in pixels (targets with indices 1--4 and 11--26) have variances up to 100 pixels, whereas the distance, orientation and dimension predictions (indices 5--10) have substantially smaller variances.

Under the heteroscedatic noise model one expects the predicted uncertainties (i.e.\ the weights used in box fitting) to behave in a meaningful way.
In Figure~\ref{fig:scatter}~(left) we show that empirically the predicted uncertainties/weights for object distance target are strongly correlated with the predicted distances (corresponding to the fact that distance estimates are more difficult at larger distances). However, Figure~\ref{fig:scatter}~(right) shows, that 2D coordinate predictions are essentially more confident with increased distance (matching the fact that 2D keypoints are better localized for distant objects).

\subsection{Object detection performance}

The metric used in this evaluation is the average precision (AP), where a detection is counted as valid if the IoU between the detection and ground truth bounding boxes is greater than 0.7 (where the IoU is calculated for bounding boxes in the image plane, in bird's-eye-view and in 3D respectively~\cite{KITTI}). We consider the 3D IoU as the most relevant detection criterion, but acknowledge that the threshold of 0.7 is very challenging for monocular methods---a depth error of only one percent is often sufficient to miss the threshold, even if all other parameter estimates are perfect. For this reason, the \textsc{KITTI} benchmark is dominated by methods using lidar data. We complement the AP metric with average orientation similarity (AOS)~\cite{KITTI} and average localization precision (ALP)~\cite{3DVP}.

We compare our method to works evaluating on the same dataset~\cite{3DVP, Mono3D, SubCNN, Deep3DBox, DeepMANTA, MonoFusion, monogrnet}. 
Some of these works were published prior to the launch of the \textsc{KITTI} 3D benchmark, and therefore used their own train-validation data split from the official training data of 7481 images. We use the same splits and call them \emph{split-1} \cite{3DOP} and \emph{split-2} \cite{3DVP}. Both validation sets \emph{split-1} and \emph{split-2} divide the training data almost in half, prevent images from the same video sequence from ending up in different sets and keep object classes balanced.
We also report results of the state-of-the-art stereo-vision based method~\cite{3DOP}, which can be seen as an optimistic baseline for this problem.
Following earlier works we focus on \texttt{Cars}, although we train on all three categories simultaneously.

Table~\ref{tab:ap3d} shows AP with the 3D IoU detection criterion for the \texttt{Cars} class. Our methods clearly outperform previous work.
Further, there is a clear ranking Method~1 $\prec$ Method~2 $\prec$ Method~3 in terms of their performance.
Table~\ref{tab:alp} similarly shows results using the ALP metric,
which is not directly connected with a 3D IoU loss (which our end-to-end model is trained on). Therefore it is not surprising that the internal ranking of our proposed methods is less pronounced.
Nevertheless, all our methods surpass previous monocular approaches and are competitive to the leading stereoscopic method.

\begin{figure}
\begin{center}
   \includegraphics[trim={0.2cm 0.7cm 15cm .5cm},clip, width=0.47\textwidth]{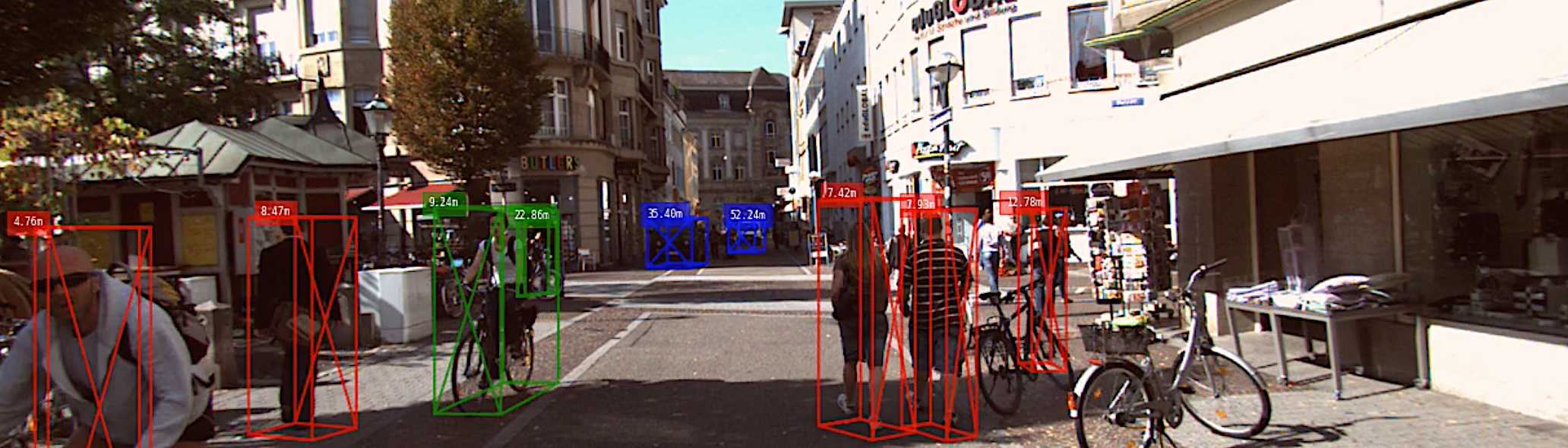}
\end{center}
   \caption{Example output projected onto image, favorably viewed at high resolution.}
     \label{fig:res1}
\end{figure}

\begin{figure}
\begin{center}
   \includegraphics[trim={0.0cm 1.0cm 0 2.0cm},clip, width=0.45\textwidth]{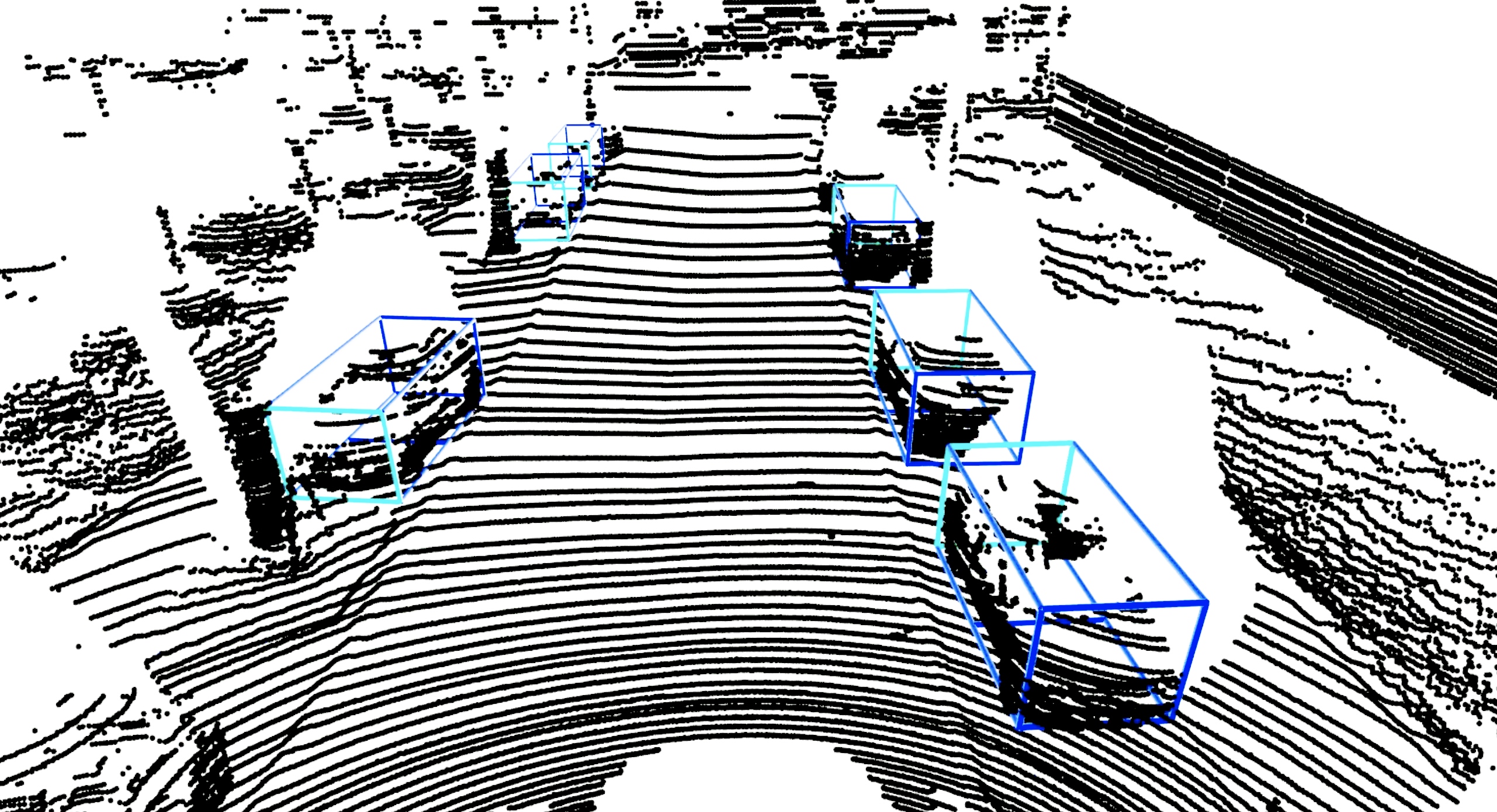}
\end{center}
   \caption{Detections visualized in the corresponding lidar point cloud. No lidar data was used during the experiments.}
     \label{fig:inLidar}
\end{figure}

We also ran inference on the \textsc{KITTI} test set and evaluated online. 
Results are shown in Table \ref{tab:online}.
Qualitative results of typical accuracy are visualized in Figures~\ref{fig:res1}~and~\ref{fig:inLidar}. 
A supplementary video\footnote{\url{https://tinyurl.com/SS3D-YouTube}} illustrates further qualitative results on the \emph{split-1} instance of the training data.

\begin{table}
\begin{center}
\scalebox{0.7}{
\begin{tabular}{|l|l||c|c|c|}
\hline
Method & Time & Easy & Moderate & Hard\\
\hline\hline
3DOP (Stereo) \cite{3DOP} & 3s &  6.55 /\quad-\quad& 5.07 / \quad-\quad & 4.10 / \quad-\quad \\
Mono3D \cite{Mono3D} & 4.2s   &  2.53 / \quad-\quad & 2.31 / \quad-\quad & 2.31 / \quad-\quad \\
Deep3DBox \cite{Deep3DBox} & N/A   & \quad-\quad / 5.85 & \quad-\quad / 4.10 & \quad-\quad / 3.84 \\
MonoFusion \cite{MonoFusion} & 0.12s   & 10.53 / 7.85 & 5.69 / 5.39 & 5.39 / 4.73\\
MonoGRNet \cite{monogrnet}  & 0.06s   & 13.88 / \quad-\quad & 10.19 / \quad-\quad & 7.62 / \quad-\quad\\
\hline
Method 1 & 0.048s & 11.54 / 8.66 & 11.07 / 7.35 & 10.12 / 5.98\\
Method 2 & 0.051s  & 13.90 / \textbf{9.55} & 12.05 / 8.07 & 11.64 / 6.99\\
Method 3 & 0.051s  & \textbf{14.52} / 9.45 & \textbf{13.15} / \textbf{8.42} & \textbf{11.85} / \textbf{7.34}\\
\hline
\end{tabular}
}
\end{center}
\caption{AP for \texttt{Cars} on data splits \emph{split-1/split-2}. Detection criterion: $IoU_{3D}\geq0.7$.}
\label{tab:ap3d}
\end{table}

\begin{table}
\begin{center}
\scalebox{0.7}{
\begin{tabular}{|l|l||c|c|c|}
\hline
Method & Time & Easy & Moderate & Hard\\
\hline\hline
3DVP \cite{3DVP} & 40s &  \quad-\quad / 45.61 &  \quad-\quad / 34.28& \quad-\quad / 27.72 \\
3DOP (Stereo) \cite{3DOP} & 3s &  \textbf{81.97} /\quad-\quad& 68.15 / \quad-\quad & 59.85 / \quad-\quad \\
Mono3D \cite{Mono3D} & 4.2s   &  48.31 / \quad-\quad & 38.98 / \quad-\quad & 34.25 / \quad-\quad \\
SubCNN \cite{SubCNN} & 2s & \quad-\quad / 39.28& \quad-\quad / 31.04 & \quad-\quad / 25.96 \\
DeepMANTA \cite{DeepMANTA} & 0.7s & 65.71 / 70.90 &  53.79 / 58.05 &  47.21 / 49.00 \\
\hline
Method 1 & 0.048s & 80.28 / \textbf{73.32} & 70.78 / 59.85 & 58.14 / 51.09\\
Method 2 & 0.051s  & 79.33 / 72.83 & \textbf{71.06} / 59.90 & 58.31 / 51.44\\
Method 3 & 0.051s & 81.22 / 72.97 & 71.05 / \textbf{59.94} & \textbf{60.22} / \textbf{51.80}\\
\hline
\end{tabular}
}
\end{center}
\caption{ALP for \texttt{Cars} on data splits \emph{split-1}/\emph{split-2}. Detection criterion: $IoU_{2D}\geq0.7$ \& 3D localization error $\leq 1m$.}
\label{tab:alp}
\end{table}

\begin{table}
\begin{center}
\scalebox{0.7}{
\begin{tabular}{|l||c|c|c|}
\hline
Metric  & Car & Pedestrian & Cyclist\\
\hline\hline
2D AP & 89.15 / 80.11 / 70.52 & 59.46 / 49.81 / 42.44 & 53.79 / 37.90 / 35.12\\
2D AOS & 89.02 / 79.70 / 69.91 & 52.70 / 43.45 / 37.20 & 44.77 / 31.17 / 28.96\\
BEV AP & 17.86 / 12.81 / 12.28 & 3.86 / 3.52 / 2.50 & 11.52 / 9.65 / 9.09\\
3D AP & 11.74 / 9.58 / 7.77 & 3.52 / 3.28 / 2.37 & 10.84 / 9.09 / 9.09\\
\hline
\end{tabular}}
\end{center}
\caption{Performance of Method~3, evaluated for all three object classes on the \textsc{KITTI} online benchmark. Easy/Moderate/Hard.}
\label{tab:online}
\end{table}

\begin{figure*}[ht]
    \begin{minipage}[b]{.325\linewidth} 
        \centering
        \includegraphics[width=0.95\textwidth]{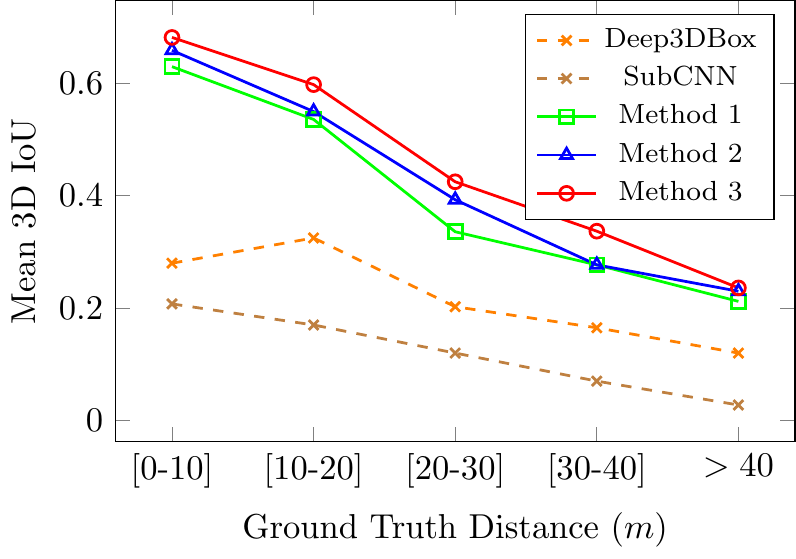}
        \label{fig:tp_a}
    \end{minipage}
    \begin{minipage}[b]{.317\linewidth} 
        \centering
        \includegraphics[width=0.95\textwidth]{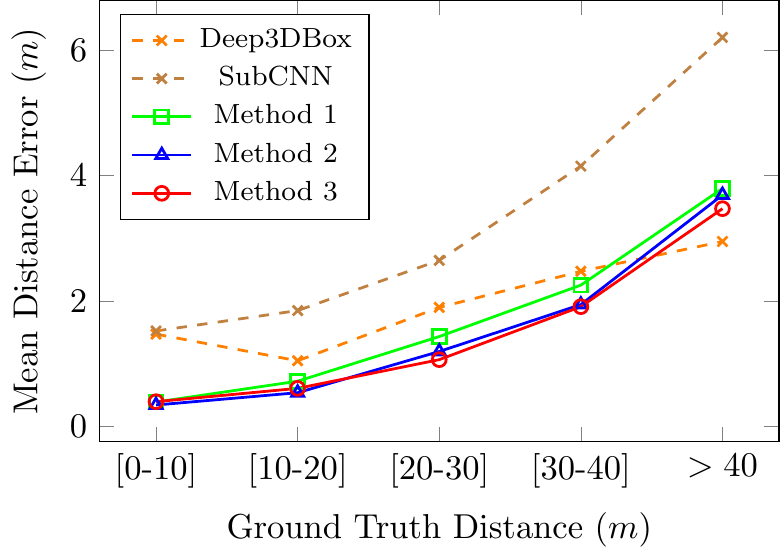}
        \label{fig:tp_b}
    \end{minipage}
    \begin{minipage}[b]{.33\linewidth} 
        \centering
        \includegraphics[width=0.95\textwidth]{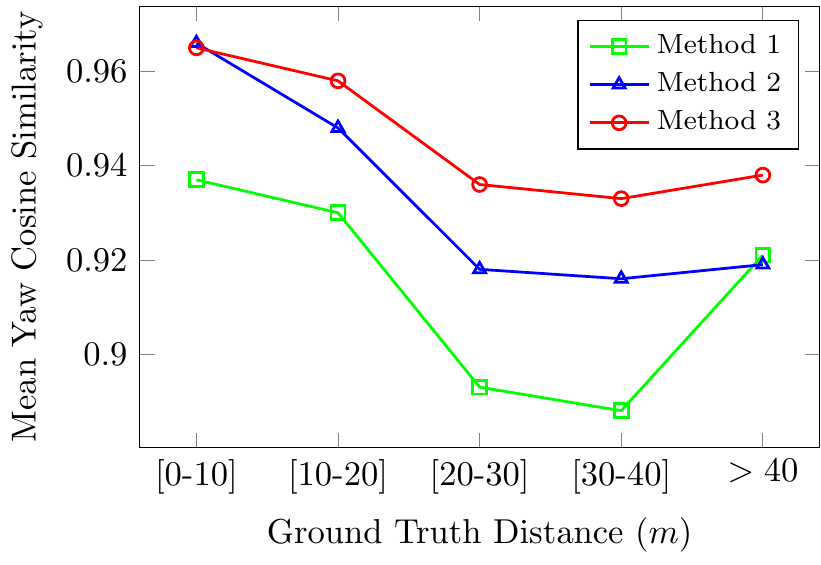}
        \label{fig:tp_c}
    \end{minipage}
    \caption{3D bounding box metrics for \texttt{Cars} on \emph{split-1} after applying a confidence threshold and assigning detections to ground truth.}
    \label{fig:tp_stats}
\end{figure*}

\subsection{3D box metrics}
While the AP metric is suitable for assessing the number of detections that match a given accuracy criterion, it does not reveal the actual accuracy of the estimated parameters. We analyze the errors of the 3D bounding box parameters by assigning each detection to its most similar ground truth box. Detections are generated by setting a class score confidence threshold for maximum $F_1$-score. We report the mean 3D IoU, the mean distance error and the mean cosine similarity, as shown in Figure~\ref{fig:tp_stats}. The methods vary more in their orientation accuracy than in their distance accuracy.
\section{Conclusion}
In this paper, we propose a joint architecture for detecting objects and fitting respective 3D bounding boxes. We demonstrate that accurate performance of monocular 3D detection can be achieved even at a high frame rate. Further, we investigate different ways of learning confidence weights for the regression outputs and demonstrate the advantage of modeling input specific confidences. Finally, we show how the complete pipeline including the CNN and the 3D bounding box optimizer can be trained end-to-end, and how this improves performance further in our application.  

For future research we intend to apply the presented framework to 3D articulated object and human pose estimation. Currently, the proposed approach works independently frame-by-frame, but leveraging temporal coherence associated with video data is another research direction.
\ificcvfinal\section*{Acknowledgements}
This work has been funded by the Swedish Research Council (grant \no 2016-04445), the Swedish Foundation
for Strategic Research (Semantic Mapping and Visual Navigation for Smart Robots), Vinnova / FFI (Perceptron,
grant \no 2017-01942 and COPPLAR, grant \no 2015-04849), the Knut and Alice Wallenberg Foundation (WASP-AI) and by Zenuity AB. It was carried out partly at Chalmers and partly at Zenuity.
\fi

{\small
\bibliographystyle{ieee}

}

\end{document}